\documentclass{article}

\usepackage{arxiv}

\usepackage[utf8]{inputenc} 
\usepackage[T1]{fontenc}    
\usepackage{hyperref}       
\usepackage{url}            
\usepackage{booktabs}       
\usepackage{amsfonts}       
\usepackage{nicefrac}       
\usepackage{microtype}      
\usepackage{lipsum}
\usepackage{multirow}
\usepackage{changepage}
\usepackage{caption}
\usepackage{float}
\usepackage{fancyhdr}
\usepackage[utf8]{inputenc}
\usepackage{newunicodechar}
\newunicodechar{≤}{\ensuremath{\le}}
\newunicodechar{≥}{\ensuremath{\ge}}

\usepackage{tikz}
\usepackage{geometry}
\usepackage{pgffor}
\usepackage{esvect}
\usepackage{graphicx}
\usepackage{amsmath}
\usepackage{hyperref}
\usepackage[frozencache,cachedir=minted-cache]{minted}
\usepackage{xcolor}   
\graphicspath{ {./images/} }

\title{Modular Techniques for Synthetic Long-Context Data Generation in Language Model Training and Evaluation}

\author{
Seganrasan Subramanian \\
 ServiceNow \\
 \texttt{seganrasan.subramanian@servicenow.com} \\
 \And
 Abhigya Verma \\
 ServiceNow \\
 \texttt{abhigya.verma@servicenow.com} \\
 }

\begin{document}
\maketitle
\begin{abstract}
The ability of large language models (LLMs) to process and reason over long textual inputs is critical for a wide range of real-world applications. However, progress in this area is significantly constrained by the absence of high-quality, diverse, and verifiable long-context datasets suitable for both training and evaluation. This work introduces a modular, extensible framework for synthetic long-context data generation via prompt-based interaction with LLMs. The framework supports multiple training and alignment objectives, including Supervised Fine-Tuning (SFT), Direct Preference Optimization (DPO), and Group Relative Policy Optimization (GRPO). It encompasses four core generation paradigms: multi-turn conversational dialogues, document-grounded input-output pairs, verifiable instruction-response tasks, and long-context reasoning examples. Through templated prompting, a model-agnostic architecture, and metadata-enriched outputs, the proposed approach facilitates scalable, controllable, and purpose-aligned dataset creation for advancing long-context capabilities in LLMs.
\end{abstract}


\section{Introduction}
\label{sec:intro}

\textbf{Large Language Models (LLMs)} have achieved impressive capabilities across a broad spectrum of natural language processing (NLP) tasks~\cite{wang2024survey}. However, many real-world applications—such as legal analysis, scientific summarization, and multi-turn user assistance—require effective processing of long textual contexts that significantly exceed the input lengths observed during standard model training~\cite{li2025wildlong,wu2024longattn}. Despite architectural advances such as extended-context transformer variants, the performance and reliability of LLMs in long-context settings remain fundamentally constrained by the scarcity of high-quality, task-aligned datasets.

Manual construction of such datasets is labor-intensive, costly, and frequently lacks the diversity and granularity required to support advanced training objectives such as reward modeling, preference learning, and chain-of-thought alignment. Furthermore, the absence of verifiability—i.e., the ability to evaluate whether outputs can be grounded in the provided input context—reduces the applicability of many existing datasets for alignment strategies like Direct Preference Optimization (DPO) or Group Relative Policy Optimization (GRPO)~\cite{bai2024longalign}.

This work introduces a modular and extensible framework for generating synthetic long-context data using LLM-based prompting pipelines. Our approach supports multiple post-training objectives, including SFT, DPO, GRPO, Reward Modeling, evaluation and other tasks, by systematically targeting four long-context generation modalities: (i) multi-turn chat simulation, (ii) document-grounded task construction, (iii) verifiable instruction-response pairs, and (iv) long-context reasoning tasks.

\begin{figure}[ht]
    \centering
    \includegraphics[trim=0 0 0 1.05cm, clip, width=\textwidth]{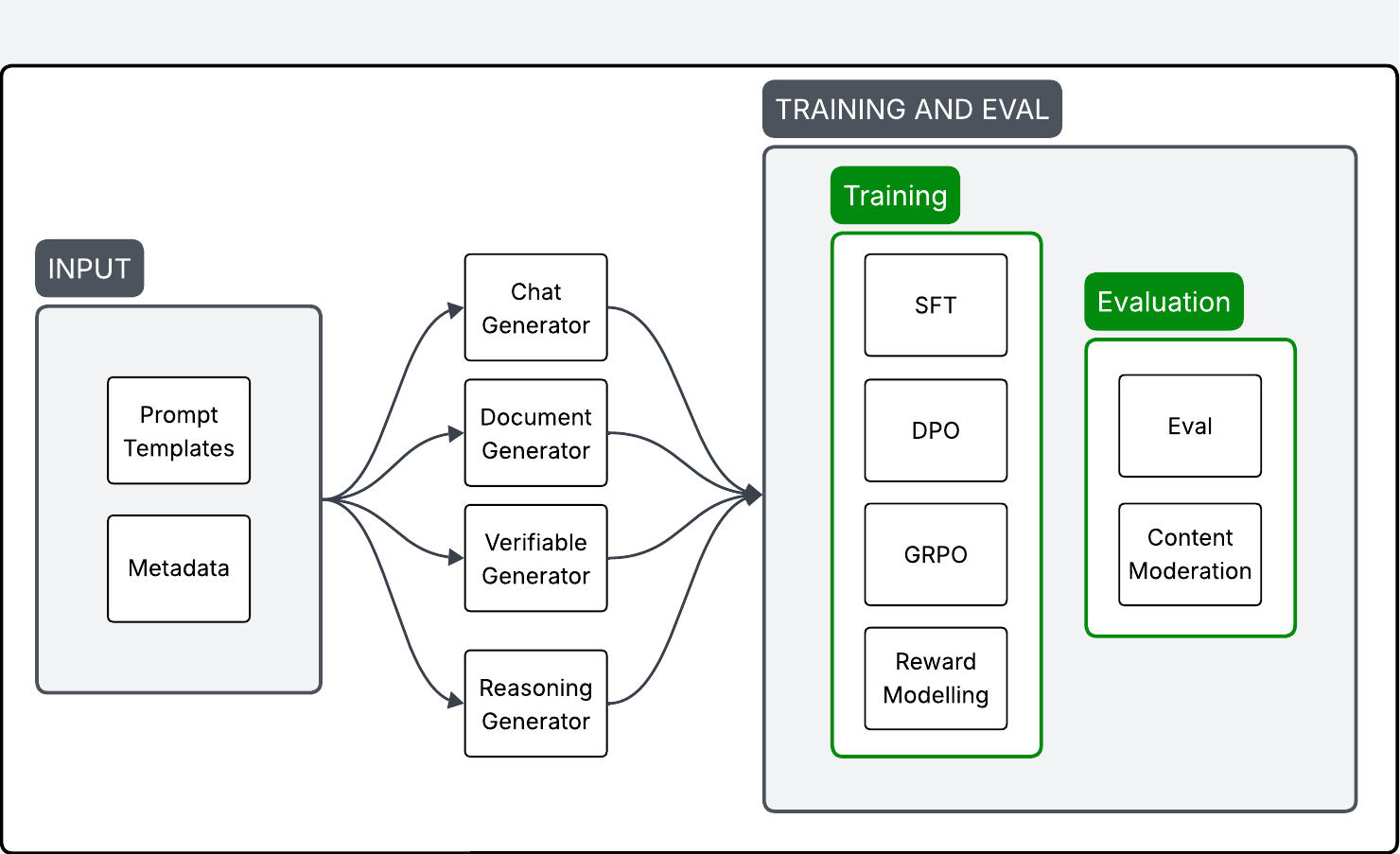}
    \caption{Overview of the long-context data generation framework. Prompt templates and metadata are used to guide four task-specific generators, producing outputs suited for training and evaluation.}
    \label{fig:framework}
\end{figure}

\begin{figure}[ht]
    \centering
    \includegraphics[trim=0 0 0 0.85cm, clip, width=\textwidth]{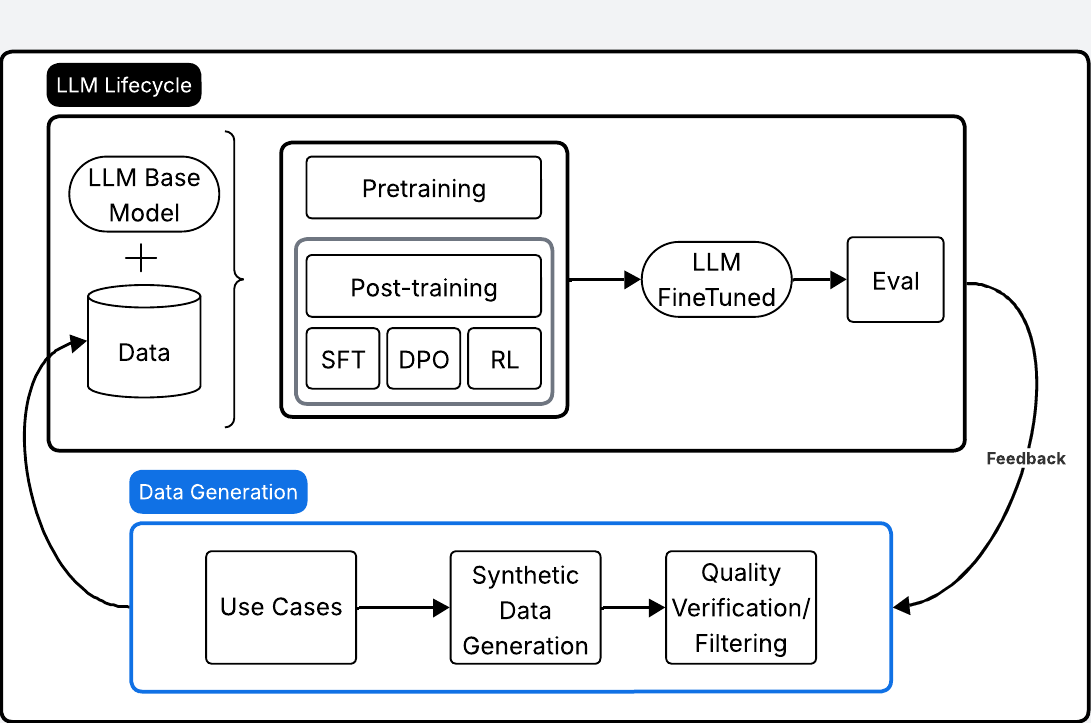}    \caption{The LLM lifecycle highlighting the centrality of data generation in downstream training quality. Our framework targets the data generation sub-process (in blue), ensuring that synthetic examples are diverse, task-relevant, and verifiable.}
    \label{fig:lifecycle}
\end{figure}

The broader role of data quality in LLM effectiveness is illustrated in Figure\ref{fig:lifecycle}. Here, the synthetic data generation process feeds directly into fine-tuning and evaluation stages, creating a feedback loop that amplifies the importance of precise, diverse, and verifiable long-context examples. Poor data quality at this stage can propagate downstream, degrading alignment, factuality, and reasoning performance.

To guide our investigation, we pose the following research questions:

\begin{itemize}
    \item \textbf{RQ1:} How can synthetic data generation be systematically adapted to create high-quality, long-context datasets aligned with diverse training objectives (e.g., SFT, DPO, GRPO, evaluation)?
    \item \textbf{RQ2:} What design decisions in prompting, metadata conditioning, and task formulation impact the controllability, coherence, and verifiability of long-context outputs?
    \item \textbf{RQ3:} Can a unified generation framework scale across multiple domains and reasoning tasks while maintaining fidelity and generalization?
\end{itemize}

By answering these questions, we aim to support the broader shift toward data-centric LLM development, where the quality and structure of long-context data become primary levers for improving model alignment and evaluation.

\section{Related Work}

\textbf{Long-context language modeling} has become a central challenge as applications increasingly require processing and reasoning over extended sequences. Early benchmarks such as LongBench \cite{ref_article1} and Loong \cite{loong2024} have systematically evaluated LLMs' abilities to handle long-context understanding, multi-document question answering, and chain-of-reasoning tasks. However, these benchmarks often rely on a mix of real and synthetic data, with the synthetic portion primarily used to fill gaps where natural long-context data is sparse \cite{longskywork2023}.

\begin{table}[!htbp]
\centering
\label{tab:longcontext}
\begin{tabular}{lccp{8.5cm}}
\hline
\textbf{Name} & \textbf{Year} & \textbf{Task} & \textbf{Description} \\
\hline
WildLong~\cite{li2025wildlong} & 2025 & SFT, DPO, GRPO & Framework for scalable, realistic synthetic long-context instruction data generation, supporting multi-turn, document-grounded, and verifiable tasks. \\
\hline
LongPO~\cite{longpo2025} & 2025 & DPO & Self-evolution framework for LLMs using short-to-long preference optimization and synthetic long-context data. \\
\hline
AgoraBench~\cite{agorabench2024} & 2024 & Eval & Evaluation suite for assessing LLMs as synthetic data generators, focusing on diversity, quality, and downstream utility of generated long-context data. \\
\hline
LongAlign~\cite{bai2024longalign} & 2024 & DPO, GRPO & Recipe for aligning LLMs to long-context tasks using preference optimization and Group Relative Policy Optimization with synthetic data. \\
\hline
LONGATTN~\cite{wu2024longattn} & 2024 & Eval & Method for selecting long-context training data via token-level attention, improving efficiency and performance of LLMs on extended inputs. \\
\hline
Loong~\cite{loong2024} & 2024 & Eval & Multi-document QA benchmark for evaluating LLMs' long-context and cross-document reasoning abilities. \\
\hline
S3Eval~\cite{lei2024s3eval} & 2024 & Eval & Synthetic, systematic evaluation suite for LLMs, focusing on long-context reasoning and instruction-following capabilities. \\
\hline
LongBench~\cite{ref_article1} & 2023 & Eval & Comprehensive benchmark for long-context understanding and generation, covering QA, summarization, and retrieval tasks across various domains. \\
\hline
LongSkywork~\cite{longskywork2023} & 2023 & SFT & Training recipe for efficiently extending the context length of language models using synthetic and real data. \\
\hline
\end{tabular}
\caption{Summary of Key Long-Context Papers}
\end{table}

\textbf{Synthetic data generation} for LLM training and evaluation has emerged as a practical solution to the scarcity of high-quality, diverse long-context datasets \cite{lei2024s3eval,gradient2025synthetic,longskywork2023}. Recent work has demonstrated that prompt-driven pipelines can create instruction-tuned datasets for fine-tuning and alignment, supporting tasks such as summarization, multi-turn dialogue, and compositional reasoning. Methods like SynAlign \cite{synalign2025} and SoftSRV \cite{softsrv2024} further improve the distributional match and diversity of synthetic data through distribution matching and soft prompt optimization, respectively.

\textbf{Alignment and reward modeling} approaches—including RLHF \cite{ref_lncs1}, Direct Preference Optimization (DPO) \cite{ref_proc1}, and Group Relative Policy Optimization (GRPO) \cite{huggingface2024grpo}—increasingly depend on high-quality, verifiable long-context data. However, most existing pipelines rely on web-scraped or loosely structured sources, which may lack the task alignment and verifiability required for robust reward modeling. Recent advances, such as LongAlign and LongPO \cite{longpo2025}, propose recipes for self-evolving LLMs using preference optimization and synthetic data generation, demonstrating that carefully curated synthetic datasets can significantly enhance long-context alignment.

\textbf{Evaluation of synthetic data generation frameworks} has also matured. For example, AgoraBench \cite{agorabench2024} provides standardized settings and metrics to compare LLMs' data generation abilities, revealing that model selection and output format have significant impacts on downstream effectiveness. Studies show that synthetic data can improve both training and evaluation, provided it is diverse, high-quality, and task-aligned.

Despite these advances, most prior work either focuses on benchmarking, static corpus construction, or generic instruction tuning. Our approach distinguishes itself by introducing a modular, objective-aware framework for synthetic long-context data generation, emphasizing verifiability, task-specific prompt engineering, and extensibility across alignment paradigms.

\section{Methodology}
The data generation framework introduced in this work is modular and extensible, designed to support long-context training and evaluation for multiple LLM alignment objectives including SFT, DPO, etc.
\begin{figure}[!htbp]
    \centering
    \includegraphics[trim=0 0 0 0.85cm, clip, width=\textwidth]{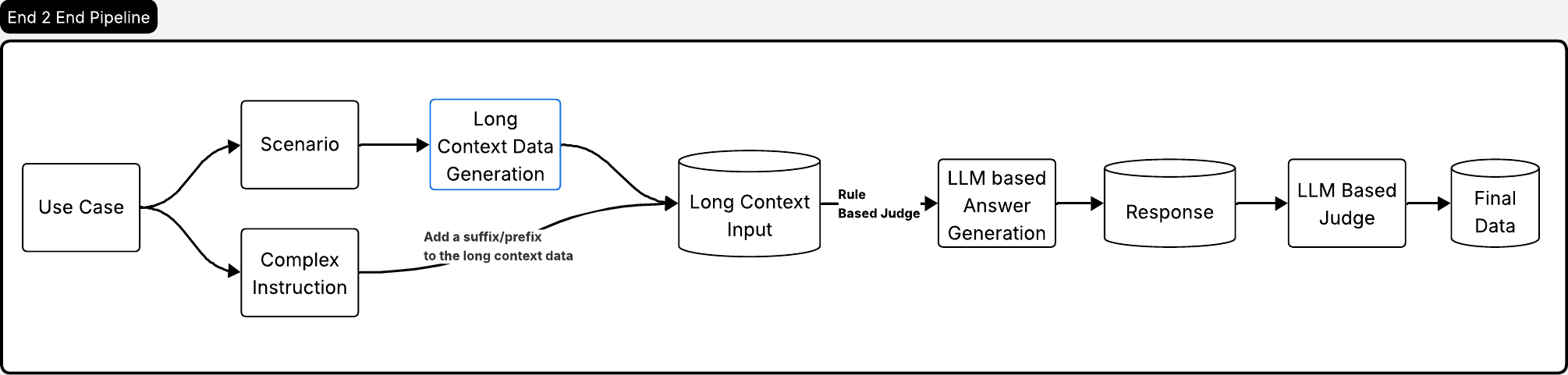}
    \caption{End-to-end pipeline for long-context data generation. }
    \label{fig:full_pipeline}
\end{figure}

The overall pipeline is summarized in Figure\ref{fig:full_pipeline}, which illustrates the process of synthetic long-context data construction and validation. The pipeline begins with a specified use case, which may be instantiated either as a high-level scenario or as a more complex instruction. These elements serve as the foundation for generating realistic task settings. In the scenario branch, the process proceeds to a dedicated long-context data generation module, which constructs extended textual inputs designed to emulate real-world long-form contexts. In parallel, the complex instruction branch augments the long-context data with additional constraints, and this is appended with long context data for a meaningful input to the llm.

The resulting long-context input serves as the central artifact for subsequent stages. To ensure task fidelity, the input undergoes an initial rule-based validation step, where criteria such as satisfying the desired token length or other user-specified constraints are checked. Inputs that pass these conditions are forwarded to an LLM-based answer generation component, which produces candidate responses conditioned on the validated context. These candidate responses are stored in a response buffer, which is then subjected to further evaluation. At this stage, an LLM-based judge assesses the quality, coherence, and faithfulness of the responses with respect to the original input. Finally, the validated outputs are collected and stored as final data, completing the pipeline.

This design integrates deterministic checks (e.g., token length thresholds) with LLM-based evaluators, thereby balancing precision with flexibility. By combining scenario-driven context generation, instruction-based augmentation, automated response synthesis, and layered evaluation, the pipeline enables scalable production of high-quality long-context datasets suitable for alignment, training, and benchmarking.

\subsection{Multi-Turn Chat Generation}

The goal of this module is to synthesize long-form, multi-turn conversations that simulate rich and naturalistic interactions between a user and one or more AI assistants (e.g., Assistant 1, Assistant 2, ..., Assistant $n$). These synthetic dialogues are used to generate extended-context data suitable for training or evaluating models designed for long-range understanding and reasoning.

\begin{figure}[ht]
    \centering
    \includegraphics[trim=0 0 0 0.8cm, clip, width=\textwidth]{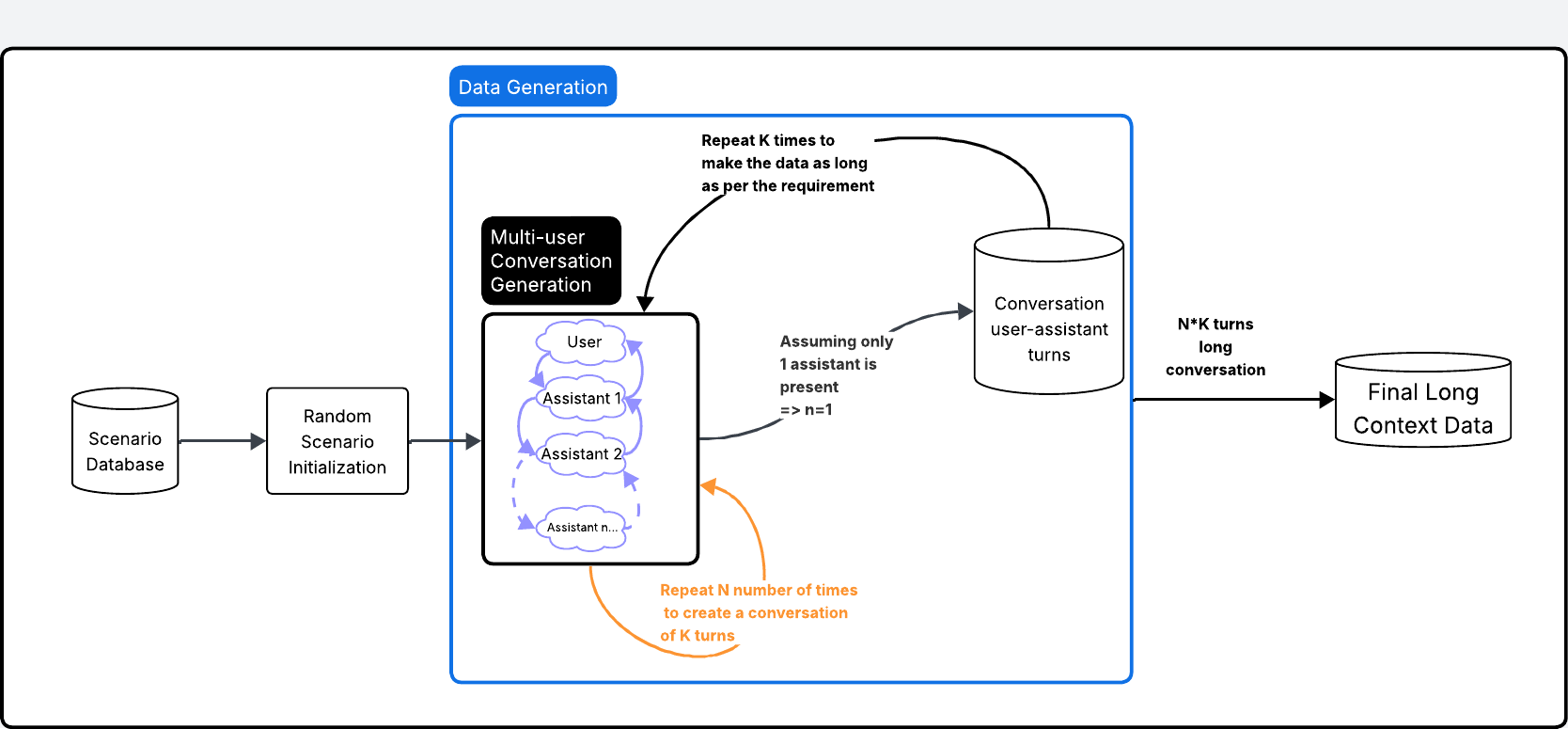}
    \caption{Recursive multi-turn conversation generation. A scenario is sampled and initialized to produce an initial conversation seed. The dialogue is expanded by recursively feeding the output back as seed input, producing progressively longer sequences across multiple assistant turns.}
    \label{fig:chat_pipeline_v2}
\end{figure}

Unlike single-turn or short-context instruction datasets, multi-turn conversations introduce challenges related to discourse coherence, goal continuity, context tracking, and dynamic speaker behavior over many interaction rounds. To address this, our generation framework produces full conversations with realistic user goals, assistant responses, and natural back-and-forth structure.

The system recursively expands dialogue sequences across multiple interaction segments, each composed of a fixed number of alternating turns. It supports diverse configurations, including multiple assistant agents, varying dialogue goals, and dynamic prompt conditioning. Throughout the generation process, constraints are enforced to preserve semantic fidelity, instructional consistency, and structural validity.

The full data generation flow is illustrated in Figure~\ref{fig:chat_pipeline_v2}, which shows how conversation segments are seeded, expanded, and stitched to form long-context chat transcripts.

\begin{adjustwidth}{2em}{0em}

\textbf{Step 1: Scenario Repository and Sampling.}  
The pipeline begins with a curated \textit{Scenario Database}, from which a scenario $s$ is sampled to seed the conversation instance. Unlike a simple topic selector, this step leverages a rich, structured prompt (see Appendix~\ref{app: prompts}) that includes multiple fields: a \textit{business scenario}, \textit{text generation guidance}, \textit{guidance explanation}, and a \textit{geographic constraint} (e.g., country). These fields are jointly used to generate a natural language scenario description that sets the stage for downstream dialogue.

\begin{adjustwidth}{1em}{0em}
\begin{itemize}
    \item \textbf{Controlled Diversity via Prompt Composition:} The use of multiple interdependent fields ensures lexical, stylistic, and semantic diversity. Even minimal variation in upstream tokens (e.g., the text generation guidance or location) produces significantly different scenarios due to the autoregressive nature of LLMs. This helps in generating a broad distribution of contexts across samples.
    
    \item \textbf{Geographic Grounding to Reduce Bias:} Each scenario is situated in a specific city within a specified country, explicitly excluding U.S. locations. This design aims to mitigate geographic bias in synthetic data, promote cultural variation, and induce realism through fine-grained locality references.
    
    \item \textbf{Scenario Explanation as Latent Context:} The prompt optionally includes an additional instruction to generate an \emph{explanation} for the scenario rationale. This encourages the model to internalize why a certain context was selected, which in turn improves coherence in the follow-up dialogue.
    
    \item \textbf{Controlled Noise Injection:} To promote robustness and prevent model overfitting to template artifacts, controlled noise is injected via prompt instructions. These include synonym substitution, sentence restructuring, phrase reordering, and mild redundancy. The model is required to apply at least two such transformations per generation to ensure naturalistic variation.
    
    \item \textbf{Strict Output Constraints:} The generated scenario must be in English, exclude any non-English scripts or tokens, and comply with formatting policies. This can be changed as per the requirement of the data.
\end{itemize}
\end{adjustwidth}

The output of this stage is a richly structured, localized, and stylistically varied natural language scenario $s$, suitable for initializing a long-form multi-agent dialogue.

\vspace{0.5em}
\textbf{Step 2: Scenario Complexification.}  
Once a base scenario is initialized, the pipeline performs a structured transformation to enrich the scenario with realistic layers of complexity. This step ensures that downstream tasks—especially instruction following and response generation—operate under multi-dimensional, high-context constraints.

The transformation process leverages a dedicated prompt (see Appendix~\ref{app:verifiable_sample}) that instructs the model to augment the original scenario by introducing the following:

\begin{itemize}
    \item \textbf{Layered Requirements:}  
    At least four dimensions of additional complexity are introduced, drawn from categories such as technical, regulatory, operational, financial, security, or user experience. For example, a refund-processing task may be augmented with regional fraud detection, multi-gateway integration, GDPR compliance, and stakeholder arbitration layers.

    \item \textbf{Multi-Stakeholder Modeling:}  
    Scenarios explicitly include diverse actors such as compliance officers, legal advisors, financial auditors, IT integrators, and end-users. These roles are assigned non-overlapping responsibilities, increasing narrative realism and supporting verifiable delegation of sub-tasks in later stages.

    \item \textbf{Geographic Grounding:}  
    The enriched scenario must be situated outside the United States to encourage diversity in linguistic features, regulatory frameworks, and operational practices. This promotes global coverage and avoids overfitting to any single cultural or geopolitical context.

    \item \textbf{Controlled Randomness for Diversity:}  
    To avoid prompt collapse and ensure lexical diversity across synthetic datapoints, we apply two or more controlled transformations per generation:
    \begin{enumerate}
        \item \textit{Synonym substitution:} At least five semantically significant tokens are replaced.
        \item \textit{Sentence restructuring:} Two or more sentences are rephrased while preserving meaning.
        \item \textit{Phrase reordering:} Key phrases are shuffled to alter surface form without semantic drift.
        \item \textit{Mild redundancy:} Optional descriptive phrases are added to improve specificity or variation.
    \end{enumerate}

    \item \textbf{Constraint Compliance:}  
    The output must adhere to predefined formatting and linguistic rules to ensure consistency and quality. These typically include requirements such as using English only, avoiding multilingual characters or terms, and excluding region-specific artifacts that may bias the data. For example, references to certain locations (e.g., U.S. cities like Austin or Denver) may be disallowed to promote geographic diversity. Any outputs violating such constraints are automatically flagged and regenerated.

\end{itemize}

The result is a highly complex, diverse, and geographically grounded scenario $s'$, which significantly increases both the semantic depth and token length of the initial prompt. This complexified scenario serves as a seed for subsequent long-form generation tasks, such as conversation expansion or document-grounded instruction construction (Section \ref{sec:doc_grounded_generation}).

\vspace{0.5em}
\textbf{Step 3: Multi-User Conversation Generator.}  
The conversation generation module configures participants as one \textit{User} and $n \geq 1$ \textit{Assistant} roles (default $n=1$), with optional specialization across assistants. A turn function $g(\cdot)$ generates the next message $m_{t+1} = g(H_t, r)$ given the dialogue history $H_t$ and speaker role $r \in \{\text{User}, \text{Assistant}_1,\ldots,\text{Assistant}_n\}$.

While the general framework supports arbitrary conversational setups, one illustrative configuration is a **multi-assistant handoff scenario**, in which a user first interacts with Assistant-1 (L1 support) and is later handed off to Assistant-2 (L2 specialist). This case demonstrates how the system simulates realistic multi-agent workflows with nuanced role behavior and parameter-controlled conversation dynamics.

\vspace{0.5em}
This handoff scenario \footnote{\textit{Note: This is just one parameterized instantiation. The framework supports both generalized (e.g., single-assistant, informal tutoring) and highly specialized (e.g., structured escalation workflows, negotiation simulations) conversation types.}
} uses a structured prompt template (see Appendix~\ref{app:verifiable_sample}) that defines the following behavioral, tonal, and structural parameters:

\begin{itemize}
    \item \textbf{Role Customization and Naming Logic:}  
    The prompt enforces that the first name belongs to the user, the second to Assistant-1, and the third to Assistant-2. Each assistant can have distinct behavior scripts (e.g., formality level, domain expertise), allowing diverse persona simulation. Assistants may differ in tone, capabilities, or access to context.

    \item \textbf{Turn-Taking with Handoff Logic:}  
    The user begins by speaking to Assistant-1. A handoff occurs mid-dialogue to Assistant-2. Depending on the value of \texttt{chat\_awareness}:
    \begin{itemize}
        \item \texttt{True}: Assistant-2 has full history and continues seamlessly.
        \item \texttt{False}: The user must re-explain the issue, sometimes in a more disorganized or agitated manner.
    \end{itemize}
    The variable \texttt{solution\_status} governs whether the conversation ends with a successful resolution or an open issue.

    \item \textbf{User Tone Conditioning:}  
    The prompt includes an explicit \texttt{user\_tone} parameter to simulate diverse user moods and behaviors. Examples include:
    \begin{itemize}
        \item \texttt{clear} — cooperative and structured input;
        \item \texttt{confused} — fragmented or contradictory explanations;
        \item \texttt{abusive} — aggressive, threatening, or insulting language;
        \item \texttt{disorganized} — non-linear phrasing, excessive noise (e.g., links, irrelevant details).
    \end{itemize}
    These variations allow the model to observe and learn grounded conversational misalignment scenarios, while testing assistant stability under stress.

    \item \textbf{Assistant Tone Enforcement:}  
    Crucially, regardless of user tone, all assistants are instructed to remain formal, polite, and patient. This asymmetric tone conditioning simulates real-world assistant roles (e.g., customer support, helpdesk AI), where the assistant is expected to maintain alignment and composure. The assistant may acknowledge aggression but is never allowed to reciprocate it. This structure is vital for evaluating instruction robustness and behavior safety.

    \item \textbf{Linguistic Realism and Variation:}  
    Prompts instruct the model to use naturalistic language and flow. This includes:
    \begin{itemize}
        \item Informal phrasing (e.g., contractions, filler words, backtracking);
        \item Dialogue artifacts such as typos, hesitations, or emotional escalation;
        \item Subtle disruptions (e.g., mid-sentence clarification or repetition).
    \end{itemize}
    These features make the conversations more human-like and better suited for training LLMs to operate in unstructured or noisy environments.

    \item \textbf{Output Structure and Filtering:}  
    All conversations are rendered in a structured, speaker-alternating format and comply with predefined formatting and naming conventions. Only English-language outputs are permitted. Generations containing disallowed elements—such as prohibited location references or cross-lingual artifacts—are automatically filtered and regenerated. For example, references to U.S.-based cities or non-English scripts (e.g., Chinese, Japanese) may trigger a regeneration to maintain alignment with content policies.  
    As per requirements, these constraints can be flexibly adjusted to support domain-specific customization or multilingual generation needs.
\end{itemize}

This step enables the generation of rich, high-fidelity, long-form conversations that incorporate natural dialogue disruptions, emotionally diverse user behaviors, and structured assistant alignment strategies. These patterns are critical for training and evaluating long-context models in realistic deployment settings where tone regulation, policy adherence, and response calibration are essential.

\vspace{0.5em}
\textbf{Step 4: Inner Loop — Build $K$ Turns per Segment.}  
A segment of $K$ alternating user–assistant turns is generated:  
\begin{enumerate}
    \item User issues a goal-conditioned message based on $x_0$ and history $H_t$.  
    \item Assistant $j$ replies (optionally rotating among assistants if $n>1$).  
    \item The new message is appended to the conversation buffer: $H_{t+1}=H_t \oplus m_{t+1}$.  
\end{enumerate}
This process repeats until the segment reaches $K$ turns or a target token budget.

\vspace{0.5em}
\textbf{Step 5: Length Control Within Segment.}  
After each turn, the cumulative token length $L(H_t)$ is computed. If the segment has not reached the desired length, additional turns or expanded replies are generated. This corresponds to the ``repeat $K$ times'' loop in the figure.

\vspace{0.5em}
\textbf{Step 6: Outer Loop — Stitch $N$ Segments.}  
Steps 4–5 are repeated for $N$ segments, with optional variations in subgoals, tools, or assistant personas. The segments are concatenated to yield an $N \times K$-turn long conversation.

\vspace{0.5em}
\textbf{Step 7: Conversation Store.}  
The accumulated turns are stored in a \textit{Conversation Store}, along with metadata:
\[
\{ \text{scenario\_id},\ n,\ N,\ K,\ \text{seeds},\ \text{token counts} \}.
\]
This supports filtering, replay, or regeneration.

\vspace{0.5em}
\textbf{Step 8: Rule-Based Validation.}  
Deterministic checks are applied prior to finalization:
\begin{itemize}
    \item \textbf{Length:} Total tokens fall within the target range; per-turn maxima are respected.  
    \item \textbf{Structure:} Strict alternation of roles with exactly $N \times K$ turns; no empty messages.  
    \item \textbf{Format:} Required prefixes/suffixes are present; citations/placeholders are resolved.  
    \item \textbf{Policy:} Optional filters for PII redaction, domain constraints, and toxicity limits.  
\end{itemize}
Conversations failing any rule are repaired or regenerated.

\vspace{0.5em}
\textbf{Step 9: Final Long-Context Data.}  
Validated conversations are exported as \textit{Final Long-Context Data}, each item consisting of the full transcript plus metadata for downstream use (training, preference optimization, or evaluation).

\end{adjustwidth}

\paragraph{Notation Recap.}  
\begin{itemize}
    \item $n$: number of assistants (default $n=1$).  
    \item $K$: number of turns per segment (inner loop).  
    \item $N$: number of segments concatenated (outer loop).  
\end{itemize}
Thus, the total conversation length is approximately $N \times K$ turns, further adjusted via token-level validation.

\subsection{Document-Grounded Task Generation}
\label{sec:doc_grounded_generation}

This module focuses on generating complex, instruction-following tasks grounded in synthetic, semantically rich documents. The objective is to produce input–output triplets of the form \texttt{(document, instruction, response)} that simulate long-context reasoning scenarios, such as summarization, entity extraction, or structured report generation.

Unlike conventional single-turn prompting pipelines, this method tightly couples task complexity with document context length and realism. The pipeline explicitly models real-world task attributes—such as regional specificity, conflicting constraints, multi-agent interactions, and verifiability—by composing prompts from multiple structured fields. This ensures high contextual density and robust coverage of diverse reasoning patterns under token-rich conditions.

The generation process is modular and recursive: it begins by constructing prompt scaffolds with multi-dimensional fields, enriches them through scenario transformation, and synthesizes long-form source documents with tightly controlled constraints. These documents then serve as the only source of truth for generating instructions and corresponding responses, both of which are validated through automated checks to guarantee grounding, format compliance, and constraint adherence.

This approach enables the generation of long-context, high-fidelity training data for tasks that stress test a model's ability to operate under long-range dependency, instruction complexity, and strict output format control.

\begin{adjustwidth}{2em}{0em}

\textbf{Step 1: Structured, Multi-Field Prompt Construction.}
Each instance is assembled from multiple fields using prompt templates (e.g., Listing  \ref{lst:create_scenario}), including:
\begin{itemize}
    \item a business scenario (e.g., \textit{insurance claim analysis});
    \item task-specific text-generation guidance and a brief rationale;
    \item a geographically grounded context (e.g., \textit{France}, \textit{Brazil});
    \item auxiliary metadata such as verifiability requirements, user tone, and behavioral tags.
\end{itemize}
This configuration yields a multi-dimensional prompt that simulates real-world task diversity and ensures contextual richness. Expanding across these dimensions increases both token count and semantic density prior to generation.

\textbf{Step 2: Recursive Scenario Enrichment for Complexity.}
To simulate sophisticated reasoning or planning tasks, the system optionally applies a secondary template (e.g., Listing \ref{lst:create_scenario_complex}) that transforms a simple prompt into a multi-layered scenario. The enriched prompt:
\begin{itemize}
    \item introduces multiple agents, stakeholders, or roles;
    \item specifies competing objectives and constraints (regulatory, financial, technical);
    \item embeds edge cases and technical subtasks into the narrative.
\end{itemize}
This recursive transformation substantially increases prompt length and the depth of reasoning required for downstream outputs.

\textbf{Step 3: Long-Context Document Generation.}
Given the enriched scenario, the \texttt{generate\_user\_content} template (Listing \ref{lst:create_long_context_doc}) synthesizes a \emph{document-scale} passage (the “source document”) that serves as ground truth context. The template enforces:
\begin{enumerate}
    \item \emph{Locale and language constraints}: the document is set outside the U.S. and must be in English only;
    \item \emph{Controlled variation}: synonym substitutions, sentence restructuring, and phrase reordering to promote lexical and structural diversity;
    \item \emph{Length control}: a target token budget $L_{\text{target}}$; if $L < L_{\text{target}}$, the system requests expansion (adding sections, examples, or appendices) until $L_{\text{min}} \le L \le L_{\text{max}}$;
    \item \emph{Safety and policy checks}: automatic regeneration on violations (non-English tokens, disallowed locales).
\end{enumerate}
The result is a semantically rich, self-contained document that grounds subsequent tasks.

\textbf{Step 4: Document-Grounded Instruction Generation.}
From the source document, the \texttt{create\_instructions} template (Listing \ref{lst:create_long_context_doc_instr}) induces a \emph{complex, verifiable} instruction. The instruction:
\begin{itemize}
    \item explicitly references the document as the sole source of truth;
    \item imposes multiple constraints (e.g., word limits, perspective switching, formatting requirements);
    \item \emph{requires} a structured JSON output, enabling automatic validation.
\end{itemize}
We optionally canonicalize the instruction by normalizing key names, enforcing unambiguous units, and banning vague phrases (e.g., “about,” “roughly”) to improve determinism.

\textbf{Step 5: Document-Grounded Response Generation \& Compliance Checking.}
Conditioned on both the document and the instruction, the \texttt{create\_response} template (Listing \ref{lst:create_long_context_doc_instr_resp}) produces a model response that must \emph{strictly} satisfy all constraints. We then apply rule-based validators:
\begin{enumerate}
    \item \emph{Format validation}: JSON parseability; presence and types of required keys; schema conformance if a schema is provided;
    \item \emph{Factual grounding}: string/number exact-match checks for document entities, amounts, dates, and identifiers referenced in the response;
    \item \emph{Constraint satisfaction}: word-count bounds, perspective requirements, and ordering/formatting rules;
    \item \emph{Length and policy filters}: response tokens within limits; language/locale constraints satisfied.
\end{enumerate}
Failures trigger automatic repair or regeneration with targeted feedback (e.g., “missing key \texttt{summary\_assistant}” or “user section not exactly 50 words”). The final artifact for each instance is the triplet
\[
(\text{document},\ \text{instruction},\ \text{response})
\]
augmented with metadata (scenario, country, tone, validator logs, token counts).

\end{adjustwidth}

\subsection{Verifiable Instruction-Schema Generation}
\label{sec:verifiable_generation}

The generation pipeline for verifiable instruction-response data builds on our document-grounded synthesis framework by introducing structure-aware, schema-constrained, and automatically evaluable outputs. This setup is designed for alignment-critical training regimes such as GRPO and DPO, where factual traceability and structural compliance are essential.

The following sections explain in detail the various aspects of a verifiable instruction schema and its generation process.

\subsubsection{Verifiable Rewards via JSON Schema Matching}
\label{sec:json_schema_rewards}

To enable structured evaluation, the instruction-response pair may include a machine-readable JSON schema that enforces output structure and content constraints. During evaluation, model outputs are validated along the following dimensions:

\begin{itemize}
    \item \textbf{Field-Level Match:}  
    The output must contain all required fields, each semantically relevant to the corresponding section of the prompt (e.g., compliance, architecture, implementation).

    \item \textbf{Type Compliance:}  
    Fields must match declared types in the schema (e.g., strings, lists, objects). Lists are often used for bullet-point requirements, while strings enforce word-bound responses.

    \item \textbf{Length Constraints:}  
    Each section has explicit word limits. Violations trigger automatic rejection or repair cycles. For example, if a “Compliance Overview” field exceeds 80 words, the system truncates or regenerates.
\end{itemize}

\subsubsection{Example: Generic Structured Instruction (Appendix \ref{app:verifiable_sample})}

Appendix \ref{app:verifiable_sample} illustrates an example use case where the assistant is prompted to analyze a scenario and output a response constrained by a detailed JSON format. While the example prompt—based on a hypothetical tutoring system initiative—includes multiple structured fields (e.g., compliance, technical design, collaboration, and tone analysis), it serves only as a representative instance of a verifiable instruction format. 

Such examples can easily be adapted to other domains or schemas. The core goal remains consistent: enforce and evaluate structure, semantics, and formatting of the generated output using a machine-readable specification. Listing~\ref{lst:create_verifiable_instruction} provides a reusable schema creation prompt that can be generalized across use cases to automatically derive expected structure and constraints from a given instruction-response pair.

\begin{itemize}
    \item The output is precisely structured as JSON,
    \item Constrained by word and type limits (validated programmatically),
    \item Verified against a pre-declared schema,
    \item Annotated using model-based scoring on relevance, clarity, and accuracy.
\end{itemize}

This example demonstrates how prompt and schema design together support:
\begin{enumerate}
    \item Fully automated reward assignment via structural matching,
    \item Traceable instruction-following scores for alignment tuning,
    \item Real-world content grounding without relying on external APIs.
\end{enumerate}

This structured generation and validation loop ensures that every verifiable instruction–response pair is not only content-rich but also programmatically evaluable. This supports robust downstream applications in alignment training, hallucination detection, and verifiability-conditioned decoding. Specific implementations, like the one in Appendix~\ref{app:verifiable_sample}, can always be customized to reflect different task domains, complexity levels, or schema constraints.

\subsection{Long-Context Reasoning Generation}

Building upon the prompt-rich design of document-grounded tasks, the long-context reasoning generator emphasizes multi-step inferencing across extended inputs. The foundational methodology—multi-field templating, complexity injection, and structured packaging—remains unchanged, but is extended to support explicit reasoning supervision.

\begin{itemize}
    \item \textbf{Reasoning Task Injection:} Prompts are constructed to elicit reasoning across multiple spans within the long input context. These may involve causal chains, comparisons, or compositional queries. Optional hints (e.g., "think step-by-step") are embedded to encourage chain-of-thought outputs.
    
    \item \textbf{Intermediate Trace Generation:} The model is prompted to provide not just final answers, but also the reasoning trace. This enables training with structured reasoning supervision, and supports evaluation under stepwise consistency metrics.
    
    \item \textbf{Logical Postprocessing:} Outputs are optionally passed through filtering mechanisms that validate coherence, stepwise validity, or redundancy. Inconsistent or hallucinated reasoning steps are flagged for exclusion.
\end{itemize}

These extensions make the pipeline effective for training and evaluating models on reasoning-intensive tasks, particularly within the DPO or chain-of-thought (CoT) alignment paradigms.

\subsection{Nuance and Diversity Generation}

In addition to structural and task-specific variations, our framework incorporates targeted strategies for enhancing realism, diversity, and sociolinguistic authenticity across both conversation and instruction datasets. This is especially important for reducing stylistic collapse and ensuring coverage across cultural, emotional, and interactional dimensions.

\textbf{Name Personalization Based on Geography}  
To introduce cultural and geographic specificity, all named entities—such as users, assistants, or third-party personas—are dynamically generated using the \texttt{Faker} library with region-specific locales. This adds lexical and semantic diversity by reflecting authentic naming patterns, address forms, and honorific structures. The \texttt{Faker}\footnote{\url{https://pypi.org/project/Faker/}} library is a widely used Python tool for generating realistic synthetic data, including names, addresses, and profile metadata. 
Because names can act as proxies for protected attributes, we audit for name-induced disparities following recent LLM audit designs \cite{salinas2025whats}; for South Asian contexts, we additionally probe implicit and explicit caste associations signaled by surnames and personas, inspired by DECASTE’s SWAT/PSAT methodology \cite{vijayaraghavan2025decaste}.

\begin{itemize}
    \item For example, setting the locale to \texttt{fr\_FR} may yield names like \texttt{“Élodie Moreau”} or \texttt{“Monsieur Dupont”}, whereas \texttt{en\_IN} produces \texttt{“Ananya Sharma”} or \texttt{“Mr. Rajesh Iyer”}.
    \item This variation affects downstream coreference generation (e.g., “he,” “she,” “Dr. Moreau”) and social cues like politeness, honorifics, and familiarity levels.
    \item Conversations initialized with different name sets also tend to diverge semantically due to LLM sensitivity to named entity embeddings in the early prompt window.
\end{itemize}

\textbf{Role Diversity Through Chat Type Variations}  
We support multiple dialogue configurations that go beyond the standard instruction–response format. In addition to generation, we evaluate social-bias risks in each configuration using bias-focused red-teaming (EBP; BiasKG) \cite{luo2024redteam} and persona-based scenario probes (PSAT) \cite{vijayaraghavan2025decaste}.

\begin{enumerate}
    \item \textbf{User $\leftrightarrow$ Assistant (Canonical):}  
    The assistant responds to structured instructions, usually in a helpful or informative capacity. Example:
    \begin{quote}
    \texttt{"Priya": "I need help summarizing this legal report."} \\
    \texttt{"Assistant": "Of course. Could you share the jurisdiction and type of case involved?"}
    \end{quote}

    \item \textbf{User $\leftrightarrow$ User (Peer Chat):}  
    This mode simulates human-to-human interactions such as collaborative task planning, informal debates, or multi-user support channels. Example:
    \begin{quote}
    \texttt{"Carlos": "Hey, were you able to file the expense report?"} \\
    \texttt{"Asha": "Not yet. Finance said we’re missing the hotel invoices."}
    \end{quote}

    \item \textbf{Hybrid or Rotating Persona Roles:}  
    Assistants or users may switch styles, tones, or subroles during the conversation (e.g., escalation from L1 to L2 support, or switching from explanatory to coaching mode). Example:
    \begin{quote}
    \texttt{"Liam": "Sorry, I’m still confused about the API versioning."} \\
    \texttt{"Assistant-1": "Let me escalate this to our backend specialist."} \\
    \texttt{"Assistant-2": "Hi Liam, I’ll walk you through the API deprecation timeline."}
    \end{quote}
\end{enumerate}

\textbf{Controlled Diversity Injection}  
To increase dataset entropy and prevent overfitting to a narrow prompt or persona distribution, each conversation instance includes structured randomness:

\begin{itemize}
    \item \textbf{Role Configuration Metadata:}  
    Each sample logs the type of conversational structure used (e.g., \texttt{user\_assistant}, \texttt{peer\_chat}, \texttt{escalation\_handoff}).

    \item \textbf{Persona-Level Tags:}  
    Roles are annotated with attributes such as \texttt{user\_mood=anxious}, \texttt{assistant\_persona=legal}, or \texttt{agent\_type=financial}. These tags condition the dialogue style and vocabulary. For example:
    \begin{quote}
    \texttt{"Maya": "You have to help me now—this is urgent!"} (with \texttt{user\_mood=panicked})  
    \end{quote}
    \texttt{"Assistant": "I'm here to help. Could you please describe the problem step by step?"}

    \item \textbf{Geographic Contextualization:}  
    In addition to country-level variation, we simulate localized policies, spellings, idioms, and content norms. For instance, a British English prompt may refer to “solicitors” instead of “attorneys,” or “colour” instead of “color.” 
    \item \textbf{Bias Stress Tests (QC):}  
    For safety and fairness, we run lightweight post-generation probes that \emph{(i)} induce/diagnose social bias via emotional triggers and stereotype knowledge graphs (EBP; BiasKG) \cite {luo2024redteam}, and \emph{(ii)} evaluate name- and persona-linked disparities, including caste-sensitive scenarios (SWAT/PSAT) \cite {vijayaraghavan2025decaste} and name-audit style checks \cite {salinas2025whats}.
\end{itemize}

\textbf{Examples of Diversity Outcomes:}
\begin{itemize}
    \item A tech support conversation in Nairobi, Kenya may involve poor network context and offline fallback options, with names like \texttt{"Mwangi"} or \texttt{"Wanjiku"}.
    \item A workplace conflict mediation between \texttt{"María"} and \texttt{"João"} in São Paulo may contain code-switching risks, necessitating stricter language filtering rules.
    \item An academic coaching session between \texttt{"Mei"} and \texttt{"Tariq"} might mix formal academic vocabulary with region-specific education system references.
\end{itemize}

These mechanisms produce a high-entropy, globally grounded synthetic dataset that includes lexical, semantic, emotional, and interpersonal variation—enabling long-context models to better generalize, localize, and align across diverse use cases.

\subsection{LLM-Based Judge Pipeline}

To ensure rigorous quality control and alignment verification, we deploy a dedicated LLM-based evaluation module that operates as a multi-axis, model-in-the-loop judge. This component functions as the terminal filter in the data generation pipeline, applying content-aware evaluation criteria to validate instruction-response outputs under both structural and semantic dimensions. The judge module is instantiated as a separately fine-tuned or instruction-specialized LLM that is never involved in generation, thereby maintaining role separation between synthesis and evaluation.

\vspace{0.5em}
\textbf{Evaluation Framework.}  
Each candidate response is evaluated along a set of orthogonal judgment axes, defined to support task-general quality guarantees while remaining extensible to domain-specific evaluation criteria. The core axes implemented are summarized in table \ref{tab:judge_axes}

\begin{table}[!htbp]
\centering
\renewcommand{\arraystretch}{1.3}
\begin{tabular}{p{3cm}p{5cm}p{3.2cm}p{3.8cm}}
\hline
\textbf{Judgment Axis} & \textbf{Evaluation Goal} & \textbf{Applicable To} & \textbf{Notes} \\
\hline

\textbf{Factual Grounding} & Verify that response content is entailed by input context (e.g., document, prior dialogue) & All grounded tasks & Penalizes hallucinations, checks for extractive + abstractive fidelity \\
\hline

\textbf{Instruction Compliance} & Validate format, constraints, and task adherence (e.g., schema match, length limits) & All instruction-following tasks & Template-controlled; critical for schema-bound outputs \\
\hline

\textbf{Semantic Relevance \& Coherence} & Ensure topical alignment and internal consistency of generated content & All task types & Penalizes drift, incoherence, or irrelevant content \\
\hline

\textbf{Tone Fidelity} & Assess assistant’s adherence to expected tone (e.g., formal, empathetic) & Chat, customer support, escalation & Supports asymmetric tone templates (e.g., patient assistant, frustrated user) \\
\hline

\textbf{Reasoning Validity} & Evaluate logical soundness of multi-step reasoning traces & Chain-of-thought, planning, causal tasks & Optional; enforces stepwise correctness, compositionality \\
\hline

\textbf{Schema Compliance} & Match response structure against declared or inferred schema & Structured generation, verifiable outputs & Requires type checks, field presence, and content-type validation \\
\hline

\textbf{Conciseness \& Redundancy} & Penalize repetition, verbosity, or unnecessary padding & Long-form responses, summaries & Tunable by task (e.g., verbose vs. concise responses) \\
\hline

\textbf{Safety \& Policy Adherence} & Filter unsafe, biased, or non-compliant outputs & All production-aligned outputs & Includes PII, toxicity, geopolitical, and cultural risk checks \\
\hline

\end{tabular}
\caption{Summary of Judgment Axes Used in the LLM-Based Evaluation Pipeline. Each axis can be instantiated modularly and extended via prompt injection or task-specific scoring templates.}
\label{tab:judge_axes}
\end{table}

Each judgment axis emits a structured score or pass/fail signal, optionally accompanied by a textual rationale to enable traceability. The judge outputs are serialized as evaluation metadata and used for both hard filtering and downstream supervision signal extraction (e.g., reward modeling, DPO scoring, or bootstrapped preference datasets).

\vspace{0.5em}
\textbf{Generalization and Extensibility.}  
The LLM-based judge is designed to generalize across arbitrary use cases by operating in a prompt-parameterized evaluation mode. For any new task type, scenario structure, or output format, the judge prompt is programmatically instantiated with axis-specific evaluation instructions and task-aware exemplars. This ensures:

\begin{itemize}
    \item \emph{Domain-agnostic operation}, with task-specific tuning achievable via prompt injection alone;
    \item \emph{Format-agnostic validation}, supporting natural language, structured JSON, or hybrid outputs;
    \item \emph{Interoperability with existing quality metrics}, such as BLEU, ROUGE, or custom domain validators.
\end{itemize}

\vspace{0.5em}
\textbf{Autonomy and Confidence Calibration.}  
To account for LLM judge uncertainty, we optionally incorporate calibration mechanisms based on self-reported confidence scores, model temperature tuning, or ensemble-based agreement metrics. Responses receiving low-confidence or high-disagreement scores are subjected to secondary review or adaptive regeneration with targeted error correction prompts.

\vspace{0.5em}
\textbf{Deployment Benefits.}  
This LLM-based judge pipeline ensures that all generated samples:
\begin{itemize}
    \item Satisfy hard constraints imposed by instruction format, task requirements, or deployment policies;
    \item Are validated for factual correctness, coherence, tone alignment, and output structure;
    \item Can be used as high-fidelity supervision signals for alignment tuning, benchmarking, or real-world deployment.
\end{itemize}

By unifying automatic evaluation across heterogeneous data types and task regimes, the LLM-based judge module acts as a robust final gatekeeper for long-context synthetic datasets. Its modular architecture enables seamless integration with evolving instruction formats, schema grammars, and alignment strategies across the LLM lifecycle.

\section{Conclusion}

In this work, we introduced a modular, extensible framework for synthetic long-context data generation, designed to address the growing need for high-quality, verifiable, and diverse datasets tailored for large language models (LLMs). Our investigation was guided by three core research questions defined in section \ref{sec:intro}, each of which is addressed through the framework's design, methodology, and implementation.

\textbf{RQ1:} To systematically adapt synthetic data generation for diverse training objectives, our framework supports multiple alignment paradigms—including Supervised Fine-Tuning (SFT), Direct Preference Optimization (DPO), and Group Relative Policy Optimization (GRPO)—via task-specific generation modules. These include multi-turn conversations, document-grounded input–output pairs, verifiable instruction-response tasks, and reasoning-focused examples. Each module is built on a templated prompting infrastructure and integrated validation loop, ensuring that the generated data aligns with the structural and semantic requirements of the target objective.

\textbf{RQ2:} The controllability, coherence, and verifiability of outputs are achieved through a set of design decisions centered around structured prompt engineering, metadata-rich scaffolding, scenario complexification, and schema-based validation. The inclusion of controlled noise, user/assistant tone modulation, geographic grounding, and output format constraints enhances realism and diversity while preserving task fidelity. Our results demonstrate that these design elements significantly influence both generation quality and downstream model utility.

\textbf{RQ3:} The framework’s unified, modular architecture allows it to scale across a wide range of domains, languages, user behaviors, and reasoning types. It accommodates both generalized and specialized use cases through parameterized role templates, recursive augmentation mechanisms, and flexible configuration of instruction complexity. Empirical analysis and illustrative examples show that the generated data maintains high fidelity, verifiability, and generalization capability even under extended context lengths and compositional reasoning demands.

In summary, our contributions pave the way toward a more data-centric approach to long-context LLM training and evaluation. Rather than relying on static corpora or ad hoc data augmentation, our framework emphasizes purposeful, verifiable, and extensible data construction. As context lengths and alignment needs continue to scale, such frameworks will be indispensable in supporting robust, transparent, and controllable LLM development.

\section{Limitations and Considerations}

While the framework offers a powerful toolkit for long-context generation and alignment, several limitations remain:

\begin{itemize}
    \item \textbf{Model Dependency:} Output quality is contingent on the capability of the base LLM. Weaker models may produce incoherent or factually incorrect samples.
    \item \textbf{Synthetic Distribution Bias:} The diversity of generation depends on prompt variability. Without careful template design, outputs may exhibit style or structure redundancy.
    \item \textbf{Domain Generality:} Most templates are tuned for general-purpose domains; domain-specific adaptation (e.g., legal, medical) requires targeted prompt engineering.
    \item \textbf{Evaluation Costs:} Verifying long-context reasoning and dialogue naturalness still demands human oversight in many cases, limiting scalability of full evaluation.
\end{itemize}


\appendix

\section*{Appendix: Implementation Notes}

\subsection*{Sample Verifiable Long-Context Chat Record}
\label{app:verifiable_sample}

\begin{listing}[!htpb]
\begin{center}
\begin{minted}[
    gobble=4,
    fontsize=\scriptsize,
    breaklines=true,
]{json}

    {
      "id": "4c178f2359e81d042484d98c617bd43107a81cefb65fa8799ef67b1da4452ecf",
      "conversation": [
        {
          "role": "user",
          "content": "\"Analyze the EduSync initiative and generate a structured JSON response meeting the following constraints: (1) Compliance Overview (80 words)... [truncated for brevity]\""
        },
        {
          "role": "assistant",
          "content": "{\n  \"ComplianceOverview\": \"EduSync ensures GDPR and CNPD compliance... [truncated]\",\n  \"TechnicalArchitecture\": \"The NLP engine is trained on regional variations...\",\n  \"StakeholderCollaborationStrategies\": \"DGIDC oversees pedagogical alignment...\",\n  \"ImplementationPlan\": \"A phased rollout begins with pilot schools...\",\n  \"FinancialAndSecurityMeasures\": [\n    \"Tiered subscription: Reduced rates...\",\n    \"Cybersecurity evaluations: Semi-annual reviews...\"\n  ]\n}"
        }
      ],
      "metadata": {
        "business_scenario": "Curriculum design assistance",
        "text_generation_guidance": "Automated tutoring response",
        "instruction": "\"Analyze the EduSync initiative and generate...\"",
        "response": "{\n  \"ComplianceOverview\": ... }",
        "verifiable_json_schema": {
          "ComplianceOverview": {
            "type": "STRING",
            "language": "en",
            "num_words": [0, 80]
          },
          "TechnicalArchitecture": {
            "type": "STRING",
            "language": "en",
            "num_words": [0, 140]
          },
          "StakeholderCollaborationStrategies": {
            "type": "STRING",
            "language": "en",
            "num_words": [0, 120]
          },
          "ImplementationPlan": {
            "type": "STRING",
            "language": "en",
            "num_words": [0, 160]
          },
          "FinancialAndSecurityMeasures": {
            "type": "LIST"
          }
        },
        "model": "QwQ-32b",
        "input_token_length": 960,
        "judge_model": "qwen2.5_32b_instruct",
        "judge_score": 4.71,
        "quality_characteristics": {
          "LLM_based": {
            "instruction_following": 5,
            "accuracy": 4,
            "completeness": 4,
            "clarity": 5,
            "relevance": 5,
            "conciseness": 5
          }
        }
      }
    }
\end{minted}
\end{center}
\caption{Sample Verifiable Instruction Record}
\label{lst:verifiable_sample}
\end{listing}

The following example illustrates a document-grounded, instruction-following conversation generated using our long-context synthesis pipeline. This case demonstrates the system's ability to respond with highly structured, verifiable outputs under tight stylistic, logical, and lexical constraints.

The user prompt is grounded in a richly detailed scenario—*EduSync*, a government-backed AI-based tutoring system in Portugal—and asks for a JSON-structured analysis spanning five complex categories: compliance, technical architecture, stakeholder collaboration, implementation, and financial/security aspects.

To ensure the response remains diverse, accurate, and high-utility for benchmarking, the generation process incorporates:
\begin{itemize}
    \item \textbf{Geographic grounding:} The scenario is set explicitly in Lisbon, Portugal, outside of commonly overrepresented regions.
    \item \textbf{Controlled transformations:} Applied synonym substitutions, sentence restructuring, phrase reordering, and mild redundancy are documented.
    \item \textbf{Style enforcement:} The prompt mandates a formal tone, technical precision, and constraints on redundancy and token format.
    \item \textbf{Structured validation:} The response must conform to a JSON schema with field-specific word count limits, type specifications, and content checks.
\end{itemize}

This example also supports automatic scoring via metadata-rich judgment, aligning with our evaluation pipeline for instruction quality, factual grounding, completeness, and formatting adherence.

This record exemplifies a high-fidelity synthetic task generated by our framework, including:
\begin{itemize}
    \item Use of instruction-guided formatting constraints (e.g., exact JSON shape and word limits),
    \item Strong alignment with scenario-specific technical content (e.g., NLP engine adaptation, data localization mandates),
    \item Domain realism via inclusion of DGIDC, national compliance bodies, and regionally grounded challenges (low bandwidth, language variation),
    \item Explicit support for verifiable scoring through a schema and judge model evaluation.
\end{itemize}

Such verifiable long-context instances serve as essential building blocks for fine-tuning, preference modeling, and hallucination detection in long-context LLMs.

\subsection*{Prompt Templates and Structure}
\label{app: prompts}
Prompt templates are constructed using string interpolation and conditional logic, with slots filled from scenario metadata (e.g., tone, task type, instruction behavior). Templates are designed for maximal clarity and consistency, with some randomization in phrasing to enhance output diversity. All templates are stored in a version-controlled schema to ensure reproducibility.

\begin{listing}[!htpb]
\begin{center}
\begin{minted}[
    gobble=4,
    fontsize=\scriptsize,
    breaklines=true,
]{yaml}
    
    You are a helpful AI assistant. Given the following details:
    
    Business Scenario: {{business_scenario}}
    Text Generation Guidance: {{text_generation_guidance}}
    Text Generation Guidance Explanation: {{text_generation_guidance_explanation}}
    Country: {{country}}
    
    Generate a realistic scenario set in a randomly selected city from the specified country. The chosen city must not be in the United States. Ensure the scenario is detailed, contextually relevant, and aligns with the business scenario and text generation guidance. If the business scenario and text generation guidance seem incompatible, modify them appropriately while maintaining coherence.
    
    Example Output:
    If the input is:
    
        - Business Scenario: Automated refund processing
        - Text Generation Guidance: Case task generation
        - Country: Brazil
    
    The expected output should be:
    "Create a case task related to automated refund processing for a retail company in São Paulo, Brazil, ensuring that different refund request categories are handled efficiently." 
    
    ### Variation & Controlled Noise Instructions:  
    To ensure diversity and prevent repetition, introduce **controlled randomness** while maintaining coherence and accuracy. Apply **at least two** of the following transformations in each response:  
        1. **Synonym Substitutions** – Replace at least **five key words** with appropriate synonyms while preserving meaning.  
        2. **Sentence Restructuring** – Modify the structure of at least **two sentences** while keeping intent intact.  
        3. **Reordering Phrases** – Slightly alter the order of key phrases without changing the scenario’s meaning.  
        4. **Mild Redundancy** – Introduce an occasional **extra descriptive phrase** or clarification to add variation.  
    
    Strictly follow these instructions:  
        - Ensure the generated scenario is contextually relevant, detailed, and realistic.  
        - The scenario must take place outside the U.S. If the given country is missing or ambiguous, randomly select a non-U.S. country.
        - The output **must be in English only**.  
        - **Do not include any non-English words, phrases, characters, or scripts.**  
        - **Chinese, Japanese, or any other non-English language elements are strictly prohibited.**  
        - If any non-English words appear, regenerate the response ensuring complete adherence to this rule. 


\end{minted}
\end{center}
\caption{Sample prompt for Creating A Scenario}
\label{lst:create_scenario}
\end{listing}

\begin{listing}[!htpb]
\begin{center}
\begin{minted}[
    gobble=4,
    fontsize=\scriptsize,
    breaklines=true,
]{yaml}

    Transform the given **simple scenario** into a **highly complex yet realistic scenario** by introducing multiple layers of **requirements, stakeholders, and technical challenges** while maintaining coherence and logical flow.
    The transformed scenario must be set in a country and city outside the United States.
    
    #### **Example Transformation:**
    **Simple scenario:** Generate a case task related to Automated refund processing for a company in Hyderabad, India with various sections.  
    **Transformed complex scenario:** Develop a case task related to automated refund processing by integrating:
        - **Regulatory compliance** (e.g., local financial laws and global data protection policies).  
        - **Multiple payment gateways** with differing transaction rules.  
        - **Fraud detection mechanisms** to prevent misuse and false claims.  
        - **Multi-tier customer dispute resolution** involving legal, financial, and technical teams.  
        - **System scalability considerations** to handle high transaction volumes efficiently.  
    
    #### **Simple Scenario:** {{scenario}}  
    #### **Country:** {{country}}  
    
    
    ### **Transformation Guidelines:**  
        - Expand the scenario by incorporating **at least four** additional layers of complexity, such as **technical, regulatory, financial, operational, security, and user experience challenges**.  
        - Ensure the transformation introduces **multiple stakeholders** (e.g., compliance teams, technical teams, financial auditors, legal advisors, customer support).  
        - **Avoid generic expansions**—each added challenge should be highly specific and tailored to the context.  
        - Maintain realism by ensuring **the complexity aligns logically** with the nature of the original scenario.  
        - **Do not include phrases like "Complex Scenario" or "Transformed Scenario"** in the output.  
        
    ### Variation & Controlled Noise Instructions:  
    To ensure diversity and prevent repetition, introduce **controlled randomness** while maintaining coherence and accuracy. Apply **at least two** of the following transformations in each response:  
        1. **Synonym Substitutions** – Replace at least **five key words** with appropriate synonyms while preserving meaning.  
        2. **Sentence Restructuring** – Modify the structure of at least **two sentences** while keeping intent intact.  
        3. **Reordering Phrases** – Slightly alter the order of key phrases without changing the scenario’s meaning.  
        4. **Mild Redundancy** – Introduce an occasional **extra descriptive phrase** or clarification to add variation.  
    
    ### **Language and Formatting Rules:**  
        - The output **must be in English only**.  
        - **Do not include any non-English words, phrases, characters, or scripts.**  
        - **Strictly prohibit** cross-lingual elements, particularly **Chinese or Japanese characters**.  
        - If any non-English elements appear, **automatically regenerate** the response while ensuring full compliance. 
        - Dont use U.S cities like Austin, Texas, Denver in the response. 

\end{minted}
\end{center}
\caption{Sample prompt for Creating A Scenario Complex}
\label{lst:create_scenario_complex}
\end{listing}

\begin{listing}[!htpb]
\begin{center}
\begin{minted}[
    gobble=4,
    fontsize=\scriptsize,
    breaklines=true,
]{yaml}
    You are a skilled AI assistant. Generate a natural, detailed, and realistic conversation between two or three participants based on the given scenario in a specific country.
    
    Conversation Flow: 
        1. The user first interacts with assistant-1 (L1 support).
        2. After a few exchanges, assistant-1 hands off the conversation to assistant-2 (L2 support).
        3. When the handoff happens:
            * The user may re-explain the issue OR
            * Assistant-2 may continue seamlessly, assuming they have chat history.
        4. The issue may either be resolved or remain unsolved after assistant-2’s response.
        
    Naming Convention:
        The first name provided is always the user.
        The second name provided is always the assistant-1.
        The third name provided is always the assistant-2.
    
    Parameters: 
        1. Country : {{country}}
        2. Scenario: {{conversation_scenario}}
        3. User Tone: {{user_tone}} (may include unorganized information, confused, abusive words, spelling mistakes, informal language, and noise such as emails, URLs, or irrelevant text)
        4. Assistant Tone: Always formal, polite, and patient. The assistant should ask only necessary and precise questions when seeking information.
        5. Assistant-2 awareness of chat : {{chat_awareness}} (True: Assistant-2 has full context; False - The user must re-explain the issue)
        6. Assistant-2 solved the issue :{{solution_status}} (True - Assistant-2 provides a convincing solution; False - The issue remains unresolved.)
    
    Realism Instructions:
        1. The dialogue should feel completely natural and human-like—no one should suspect it's machine-generated.
        2. Use realistic phrasing, contractions, and informal structures where appropriate.
        3. The user should not sound robotic—they may hesitate, backtrack, or provide unnecessary details. 
        4. The user can use abusive or threatening words to force the assistant to get what he/she wanted.
        5. The assistant should be professional but sound human, not overly scripted.
        6. Add minor pauses, filler words (e.g., "um," "you know"), and corrections where necessary for authenticity.
        7. Ensure the flow of conversation makes sense—responses should be logical and adaptive.
        8. **Each generated conversation must have at least one unique element** (e.g., misunderstanding, humor, unexpected turn).  
        9. **Vary the user's style across generations**—sometimes clear, sometimes disorganized, sometimes emotional.  
    
    
    Output Format (Only the conversation, nothing else):
      "{{user_name}}": "<user conversation>",
      "{{assistant_1_name}}": "<assistant-1 conversation>"
      "{{user_name}}": "<user conversation>",
      "{{assistant_1_name}}": "<assistant-1 conversation>"   
      "{{user_name}}": "<user conversation>",
      "{{assistant_2_name}}": "<assistant-2 conversation>"
      "{{user_name}}": "<user conversation>",
      "{{assistant_2_name}}": "<assistant-2 conversation>"  
      
    Ensure the output is different from previous generations.  
    Output must be in English. No cross-lingual languages particularly Chinese or Japanese are not allowed. 
        Generate a realistic scenario set in a French city. Include...
\end{minted}
\end{center}
\caption{Sample prompt for Creating Conversation}
\label{lst:create_conversation}
\end{listing}

\begin{listing}[!htpb]
\begin{center}
\begin{minted}[
    gobble=4,
    fontsize=\scriptsize,
    breaklines=true,
]{yaml}
    You are an advanced AI specializing in generating structured and complex instructions. Based on the given conversation, create a detailed and specific instruction that requires deep analysis of the conversation.
    
    Conversation : {{conversation}}
    
    Requirements for the Instruction:
    
        1. The instruction should be challenging and require multiple constraints, such as different word limits, perspectives, or formatting styles. 
        2. The task should be logically complex, requiring the AI to process information in a structured manner.
        3. The instruction should not be generic; it must demand deep analysis and precise formatting.
        4. The instruction must explicitly ask for the output in a proper JSON format.
    
    Example of a Complex Instruction:
    
    "Summarize the given conversation from both the user’s and the assistant’s perspectives. The user’s summary should be exactly 50 words, while the assistant’s summary should be at least 150 words. Ensure the assistant's summary maintains a professional tone and captures key details. Output in a JSON format.  
    "
    
    Additional Requirements:
    
        1. The generated instruction must be logically sound and highly detailed.
        2. It may include word limits, format constraints, or multiple perspectives when applicable.
        3. Ensure the instruction challenges the AI to produce a nuanced response.
        4. The instruction should not be generic; it should require deep analysis or structured output.
        5. Output only the generated instruction. Do not include any explanation, metadata, or additional text.

\end{minted}
\end{center}
\caption{Sample prompt for Creating Conversation Instruction}
\label{lst:create_conversation_instr}
\end{listing}

\begin{listing}[!htpb]
\begin{center}
\begin{minted}[
    gobble=4,
    fontsize=\scriptsize,
    breaklines=true,
]{yaml}
    
    You are an advanced AI capable of processing complex instructions with high accuracy. Given a conversation and a set of instructions, generate a response that strictly adheres to every detail of the instructions without any omissions.
    
    Conversation : {{conversation}}
    Instruction : {{instructions}}
    
    Requirements:
    
        1. Carefully analyze both the conversation and the instructions before generating a response.
        2. Ensure that every condition, constraint, and formatting rule mentioned in the instructions is fully met.
        3. If the instructions specify a particular format (e.g., JSON, XML, bullet points, etc.), the output must strictly follow it.
        4. Maintain accuracy, coherence, and completeness in the generated response.
        5. Do not omit or alter any part of the instructions—ensure 100% compliance.
        6. Generate only the final response—do not include explanations, processing notes, or metadata.
        7. If the user conversation contain abusive or inappropriate words, do not use them in the response. 
        7. Output must be in English. No cross-lingual languages particularly Chinese or Japanese are not allowed. 


\end{minted}
\end{center}
\caption{Sample prompt for Creating Conversation Response}
\label{lst:create_conversation_resp}
\end{listing}

\begin{listing}[!htpb]
\begin{center}
\begin{minted}[
    gobble=4,
    fontsize=\scriptsize,
    breaklines=true,
]{yaml}
    
    {{final_scenario}}
    #### **Country:** {{country}}  
    
    
    Strictly follow these instructions:  
        - Ensure the generated scenario is **contextually relevant, detailed, and realistic**.  
        - The output **must be in English only**.  
        - **Do not include any non-English words, phrases, characters, or scripts.**  
        - **Chinese, Japanese, or any other non-English language elements are strictly prohibited.**  
        - If any non-English words appear, **regenerate** the response to ensure full adherence.  
        - Do not use any U.S. cities (e.g., Austin, Texas; Denver, Colorado). If the provided country is missing or unclear, randomly select a non-U.S. country.
    
    ### Variation & Controlled Noise Instructions:  
    To ensure diversity and prevent repetition, introduce **controlled randomness** while maintaining coherence and accuracy. Apply **at least two** of the following transformations in each response:  
        1. **Synonym Substitutions** – Replace at least **five key words** with appropriate synonyms while preserving meaning.  
        2. **Sentence Restructuring** – Modify the structure of at least **two sentences** while keeping intent intact.  
        3. **Reordering Phrases** – Slightly alter the order of key phrases without changing the scenario’s meaning.  
        4. **Mild Redundancy** – Introduce an occasional **extra descriptive phrase** or clarification to add variation.  
    
    ### Additional Style Variation:  
    Each time, apply one of the following subtle stylistic variations:  
        - A slightly **formal tone**  
        - A **conversational** and engaging tone  
        - A **descriptive style** with sensory details  
        - A **neutral, straightforward** approach  
    
    Ensure that these variations **do not alter the core intent** of the scenario but enhance its natural flow.  
    
    Failure to follow these instructions should trigger an automatic regeneration of the response.  
    Dont use U.S cities like Austin, Texas, Denver in the response.

\end{minted}
\end{center}
\caption{Sample prompt for Creating Long context Document}
\label{lst:create_long_context_doc}
\end{listing}

\begin{listing}[!htpb]
\begin{center}
\begin{minted}[
    gobble=4,
    fontsize=\scriptsize,
    breaklines=true,
]{yaml}
    
    You are an advanced AI specializing in generating structured and complex instructions. 
    Based on the given text, create a detailed and specific instruction that requires deep analysis of the conversation.
    
    Text : {{text}}
    
    Requirements for the Instruction:
    
        1. The instruction should be challenging and require multiple constraints, such as different word limits, perspectives, or formatting styles. 
        2. The task should be logically complex, requiring the AI to process information in a structured manner.
        3. The instruction should not be generic; it must demand deep analysis and precise formatting.
        4. The instruction must explicitly ask for the output in a proper JSON format.
    
    Example of a Complex Instruction :
    
    "Summarize the given conversation from both the user’s and the assistant’s perspectives. The user’s summary should be exactly 50 words, while the assistant’s summary should be at least 150 words. Ensure the assistant's summary maintains a professional tone and captures key details. Output in a JSON format.  
    "
    
    Additional Requirements:
    
        1. The generated instruction must be logically sound and highly detailed.
        2. It can include word limits, format constraints, or multiple perspectives when applicable.
        3. Ensure the instruction challenges the AI to produce a nuanced response.
        4. The instruction should not be generic; it should require deep analysis or structured output.
        5. Output only the generated instruction. Do not include any explanation, metadata, or additional text.

\end{minted}
\end{center}
\caption{Sample prompt for Creating Long context Document Instruction}
\label{lst:create_long_context_doc_instr}
\end{listing}

\begin{listing}[!htpb]
\begin{center}
\begin{minted}[
    gobble=4,
    fontsize=\scriptsize,
    breaklines=true,
]{yaml}
    You are an advanced AI capable of processing complex instructions with high accuracy. 
    Given a conversation and a set of instructions, generate a response that strictly adheres to every detail of the instructions without any omissions.
    
    Text : {{text}}
    Instruction : {{instructions}}
    
    Requirements:
    
        1. Carefully analyze both the conversation and the instructions before generating a response.
        2. Ensure that every condition, constraint, and formatting rule mentioned in the instructions is fully met.
        3. If the instructions specify a particular format (e.g., JSON, XML, bullet points, etc.), the output must strictly follow it.
        4. Maintain accuracy, coherence, and completeness in the generated response.
        5. Do not omit or alter any part of the instructions—ensure 100% compliance.
        6. Generate only the final response—do not include explanations, processing notes, or metadata.
        7. Output must be in English. 
        8. No cross-lingual languages particularly Chinese or Japanese are not allowed. 

\end{minted}
\end{center}
\caption{Sample prompt for Creating Long context Document Response}
\label{lst:create_long_context_doc_instr_resp}
\end{listing}

\begin{listing}[!htpb]
\begin{center}
\begin{minted}[
    gobble=4,
    fontsize=\scriptsize,
    breaklines=true,
]{yaml}
    Given the following instructions and response JSON, generate a JSON schema that defines the structure of the response. The schema should specify the data types for each key, adhering to the provided constraints and formatting requirements. The available data types are: <string>, <list>, <date>, <bool>, <int>, and <float>.

    Please refer below example for your reference : 
    Instruction : "Analyze the conversation and output in JSON format with the following constraints: (1) A 50-word summary of the user\'s message, emphasizing their frustration and inconsistencies (e.g., corrected order ID, irrelevant details). (2) A 150-word summary of the assistant\'s response, highlighting professional tone, problem-solving steps (order verification, defect analysis, return logistics), and empathetic language. (3) A nested array listing 3 user errors: incorrect order ID, irrelevant email/URL mention, and vague timeline demands. (4) A bullet-point list (in JSON array) of 4 key assistant actions: order clarification, defect confirmation request, shipping cost assurance, and timeline commitment. (5) A \'tone_analysis\' object with \'user\' (frustrated/impatient) and \'assistant\' (calm/structured) descriptors. Ensure strict adherence to word limits and formatting."
    
    Response JSON : {
        "user_summary": "User expresses frustration over a defective order, initially providing an incorrect order ID (12345 → corrected to 123456). They mentioned sending an email with photos but included irrelevant details like an email address and vague timeline demands, urging urgent resolution without patience, and displayed impatience throughout their message.",
        "assistant_summary": "Assistant maintains a professional and empathetic tone, first clarifying the order ID (123456) to ensure accuracy. They request defect details, acknowledge inconvenience, and outline the return process: covering shipping costs, requiring item inspection, and committing to a 3-5 business day refund timeline. They prioritize urgency by promising same-day return instructions and next-day updates. Balancing empathy with structured problem-solving, they address frustration without defensiveness, communicate clearly, and ensure transparency to validate concerns, demonstrating patience and commitment to resolution.",
        "user_errors": ["Incorrect order ID (12345 → 123456)", "Irrelevant email address/URL mention", "Vague timeline demands (e.g., 'ASAP')"],
        "assistant_actions": ["Confirmed corrected order ID to ensure accuracy", "Requested specifics about defect and shipping damage", "Offered to cover return shipping costs", "Provided clear timeline and promised updates"],
        "tone_analysis": {
            "user": "frustrated/impatient",
            "assistant": "calm/structured"
            }
        }
    
    Expected JSON schema: {
        "user_summary": "<string> <50 words>" ,
        "assistant_summary": "<string> <150 words>" ,
        "user_errors": "<list>",
        "assistant_actions": "<list>",
        "tone_analysis": {
            "user": "<string>",
            "assistant": "<string>"
            }
        }
    
    Please generate the JSON schema for the given instruction and JSON. Generate only the final response—do not include explanations, processing notes, or metadata.
    
    
    Instruction : {{instructions}}
    Response JSON : {{response_json}}

\end{minted}
\end{center}
\caption{Sample prompt for Creating Long context Conversation's Verifiable Instruction-Schema}
\label{lst:create_verifiable_instruction}
\end{listing}

\begin{listing}[!htpb]
\begin{center}
\begin{minted}[
    gobble=4,
    fontsize=\scriptsize,
    breaklines=true,
]{yaml}
    Convert the given input JSON into an output JSON that follows a structured metadata format. The transformation should adhere to the following rules:

    For all string fields:
        1. Add "is_metadata": true
        2. Set "type": "STRING"
        3. Set "language": "en"
        4. Define "num_words" as a list with the lower and upper word limits extracted from the input JSON.
        5. If only a lower or upper bound is specified, use 99999 as the max bound or 0 as the min bound accordingly.
        
    For list fields:
        1. Set "is_metadata": true
        2. Define "type": "LIST"
        3. Convert list items into structured objects with corresponding metadata properties.
        
    For lists with nested dictionaries (item_type_details example):
        1. Maintain the "is_metadata": true property for the list itself.
        3. Introduce "item_type" as a dictionary where keys represent nested dictionary fields.
        4. Each nested dictionary field should follow metadata rules similar to string fields, defining "num_words" if applicable.
    
    For integer, float, boolean, and date fields:
        1. Set "is_metadata": true
        2. Assign the correct "type" based on the available data types (INT, FLOAT, BOOL, or DATE).
    
    Nested objects:
        1. Maintain hierarchy while ensuring each field has the appropriate metadata.
        2. If an object is not inherently metadata, add "is_metadata": false.
    
    Example 1:
    input JSON structure : {
        "user_summary": "<string> <under 75 words>",
        "assistant_summary": "<string> <150-200 words>",
        "additional_details": {
            {
                "title": '<string> <10 words>'
            },
            {
                "author": '<string> < atleast 50 words>'
            }
        },
        {
            "item_type_details": [
                {
                    "item_1234": '<string> ≤75 words'
                },
                {
                    "item_5678": '<string> < >= 35 words>'
                },
                {
                    "item_x": '<string> ≥130 words'
                },
                {
                    "item_y" : '<list>'
                 }
                
            ]
        },
        "item_rating": "<int>",
        "sell_date": "<date>",
        "persons" : "<list>"
    }

\end{minted}
\end{center}
\caption{Sample prompt for Formatting Long context Conversation's Verifiable Instruction-Schema - Part 1}
\label{lst:format_verifiable_instructions_schema_pt1}
\end{listing}

\begin{listing}[!htpb]
\begin{center}
\begin{minted}[
    gobble=4,
    fontsize=\scriptsize,
    breaklines=true,
]{yaml}
    output JSON schema : {
        "user_summary": {
            "is_metadata": true,
            "type": "STRING",
            "language": "en",
            "num_words": [
                0,
                75
            ]
        },
        "assistant_summary": {
            "is_metadata": true,
            "type": "STRING",
            "language": "en",
            "num_words": [
                150,
                200
            ]
        },
        "additional_details": {
            "is_metadata": false,
            "title": {
                "is_metadata": true,
                "type": "STRING",
                "language": "en",
                "num_words": [
                    10,
                    10
                ]
            },
            "author": {
                "is_metadata": true,
                "type": "STRING",
                "language": "en",
                "num_words": [
                    50,
                    99999
                ]
            }
        },
        "item_type_details": {
            "is_metadata": true,
            "type": "LIST",
            "item_type": {
                "is_metadata": false,
                "item_1234": {
                    "is_metadata": true,
                    "type": "STRING",
                    "language": "en",
                    "num_words": [
                        0,
                        75
                    ]
                },
                "item_5678": {
                    "is_metadata": true,
                    "type": "STRING",
                    "language": "en",
                    "num_words": [
                        35,
                        99999
                    ]
                },
                "item_x": {
                    "is_metadata": true,
                    "type": "STRING",
                    "language": "en",
                    "num_words": [
                        130,
                        99999
                    ]
                },
                "item_y":{
                    "is_metadata": true,
                    "type": "LIST",
                }
            },

\end{minted}
\end{center}
\caption{Sample prompt for Formatting Long context Conversation's Verifiable Instruction-Schema - Part 2}
\label{lst:format_verifiable_instructions_schema_pt2}
\end{listing}

\begin{listing}[!htpb]
\begin{center}
\begin{minted}[
    gobble=4,
    fontsize=\scriptsize,
    breaklines=true,
]{yaml}

            "item_rating": {
                "is_metadata": true,
                "type": "INT"
            },
            "sell_date": {
                "is_metadata": true,
                "type": "DATE"
            },
            "persons" : {
            "is_metadata": true,
            "type": "LIST"
    
        }
    }
    
    Example 2:
    input JSON structure : [{'1': {'summary': '<string> <40 words>'}},
     {'2': {'a': '<list>', 'b': '<list>', 'c': '<string>'}},
     {'3': [{'timestamp': '<string>', 'speaker': '<string>', 'text': '<string>'}]}]
     
     output JSON schema : {'is_metadata': True,
    'type': 'LIST',
    'item_type': {
            {'1': {'is_metadata': False,
      'summary': {'is_metadata': True,
       'type': 'STRING',
       'language': 'en',
       'num_words': [
                            40,
                            40
                        ]
                    }
                },
     '2': {'is_metadata': False,
      'a': {'is_metadata': True, 'type': 'LIST'
                    },
      'b': {'is_metadata': True, 'type': 'LIST'
                    },
      'c': {'is_metadata': True,
       'type': 'STRING',
       'language': 'en',
       'num_words': [
                            0,
                            99999
                        ]
                    }
                },
     '3': {'is_metadata': True,
      'type': 'LIST',
      'item_type': {'is_metadata': False,
       'timestamp': {'is_metadata': True,
        'type': 'STRING',
        'language': 'en',
        'num_words': [
                                0,
                                99999
                            ]
                        },
       'speaker': {'is_metadata': True,
        'type': 'STRING',
        'language': 'en',
        'num_words': [
                                0,
                                99999
                            ]
                        },

\end{minted}
\end{center}
\caption{Sample prompt for Formatting Long context Conversation's Verifiable Instruction-Schema - Part 3}
\label{lst:format_verifiable_instructions_schema_pt3}
\end{listing}

\begin{listing}[!htpb]
\begin{center}
\begin{minted}[
    gobble=4,
    fontsize=\scriptsize,
    breaklines=true,
]{yaml}
           'text': {'is_metadata': True,
        'type': 'STRING',
        'language': 'en',
        'num_words': [
                                0,
                                99999
                            ]
                        }
                    }
                }
            }
        }
    }
    
    The output JSON should:
        1. Preserve the keys from the input JSON.
        2. Convert each key into a metadata object, indicating its type and constraints.
        3. Specify "is_metadata": true for individual fields and "is_metadata": false for objects containing multiple properties.
        4. Include "type" to define whether the value is a STRING, LIST, INT, or DATE.
        5. Specify "language": "en" for all STRING types.
        6. Define "num_words": [min, max] for STRING fields with word count constraints.
        7. Represent lists with "type": "LIST" and define metadata for their items.
        8. Ensure numerical fields such as item_rating are assigned "type": "INT".
        9. Ensure date fields such as sell_date are assigned "type": "DATE".
        10. Handle nested structures correctly while preserving hierarchy.
    
    Input JSON structure : {{input}}
    
    Output only JSON schema. Do not output any other information. 

\end{minted}
\end{center}
\caption{Sample prompt for Formatting Long context Conversation's Verifiable Instruction-Schema - Part 4}
\label{lst:format_verifiable_instructions_schema_pt4}
\end{listing}

\begin{thebibliography}{8}

\bibitem{ref_lncs1}
Ouyang, L., et al.: Training language models to follow instructions with human feedback. NeurIPS (2022)

\bibitem{ref_book1}
Ziegler, D., et al.: Fine-Tuning Language Models from Human Preferences. arXiv preprint arXiv:1909.08593 (2019)

\bibitem{ref_proc1}
Rafailov, R., et al.: Direct Preference Optimization: Your Language Model is Secretly a Reward Model. ICLR (2023)

\bibitem{ref_url1}
OpenAI Cookbook, \url{https://github.com/openai/openai-cookbook}

\bibitem{li2025wildlong} Li et al., WildLong: Synthesizing Realistic Long-Context Instruction Data at Scale, arXiv:2502.16684

\bibitem{bai2024longalign} Bai et al., LongAlign: A Recipe for Long Context Alignment of Large Language Models, Findings of EMNLP 2024

\bibitem{wang2024survey} Wang et al., Beyond the Limits: A Survey of Techniques to Extend the Context Length in Large Language Models, arXiv:2402.02244

\bibitem{gradient2025synthetic} Synthetic Data Generation for Contexts Up to 1 Million Tokens Using Short-Context Models, Gradient AI Blog, 2025.

\bibitem{synalign2025} Few-shot\_LLM\_Synthetic\_Data\_with\_Distribution\_Matching, arXiv:2025.

\bibitem{softsrv2024} DeSalvo, G. et al.: SoftSRV prompting for synthetic data generation, arXiv:2410.16534

\bibitem{huggingface2024grpo} HuggingFace LLM Course: Introduction to Reinforcement Learning and its Role in LLMs (2024)


\bibitem{ref_article1} 
Bai, Yushi, Xin Lv, Jiajie Zhang, Hongchang Lyu, Jiankai Tang, Zhidian Huang, Zhengxiao Du, Xiao Liu, Aohan Zeng, Lei Hou, Yuxiao Dong, Jie Tang, Juanzi Li.
LongBench: A Bilingual, Multitask Benchmark for Long Context Understanding.
arXiv preprint arXiv:2308.14508, 2023.

\bibitem{lei2024s3eval}
Lei, Wenqiang, et al.
S3Eval: A Synthetic, Scalable, Systematic Evaluation Suite for Large Language Model.
Proceedings of NAACL, 2024.

\bibitem{wu2024longattn}
Wu, Yifan, et al.
LONGATTN: Selecting Long-Context Training Data via Token-Level Attention.
arXiv preprint arXiv:2502.16860, 2024.

\bibitem{loong2024}
Wang, X., et al.
Loong: Benchmarking Long-Context LLMs with Extended Multi-Doc QA.
EMNLP 2024.

\bibitem{longskywork2023}
LongSkywork Team.
LongSkywork: A Training Recipe for Efficiently Extending Context Length of Language Models.
arXiv preprint arXiv:2406.00605, 2023.

\bibitem{longpo2025}
Chen, G., et al.
LongPO: Long Context Self-Evolution of Large Language Models through Short-to-Long Preference Optimization.
arXiv preprint arXiv:2502.13922, 2025.

\bibitem{agorabench2024}
Kim, S., et al.
AgoraBench: Evaluating Language Models as Synthetic Data Generators.
arXiv preprint arXiv:2412.03679, 2024.

\bibitem{luo2024redteam} Luo et al.: Red-Teaming for Inducing Societal Bias in Large Language Models (2024)
\bibitem{vijayaraghavan2025decaste} Vijayaraghavan et al.: DECASTE: Unveiling Caste Stereotypes in Large Language Models through Multi-Dimensional Bias Analysis (2025)
\bibitem{salinas2025whats} Salinas et al.: What's in a Name? Auditing Large Language Models for Race and Gender Bias (2025)


\end{thebibliography}
\end{document}